# Deception Detection in Videos using the Facial Action Coding System

**Author Information:**

1. **Hammad-Ud-Din Ahmed**

   **Email: connecthammad@gmail.com**

2. **Usama Ijaz Bajwa (Corresponding Author)**

   **Email: usamabajwa@cuilahore.edu.pk**

   **ORCID: 0000-0001-5755-1194**

3. **Fan Zhang**

   **Email: fzhang@us.ibm.com**

   **ORCID:  0000-0002-5576-7271**

4. **Muhammad Waqas Anwar**

   **Email: waqasanwar@cuilahore.edu.pk**

   **ORCID: 0000-0002-7822-8983**

# Deception Detection in Videos using the Facial Action Coding System

*Abstract*—Facts are important in decision making in every situation, which is why it is important to catch deceptive information before they are accepted as facts. Deception detection in videos has gained traction in recent times for its various real-life application. In our approach, we extract facial action units using the facial action coding system which we use as parameters for training a deep learning model. We specifically use long short-term memory (LSTM) which we trained using the real-life trial dataset and it provided one of the best facial only approaches to deception detection. We also tested cross-dataset validation using the Real-life trial dataset, the Silesian Deception Dataset, and the Bag-of-lies Deception Dataset which has not yet been attempted by anyone else for a deception detection system. We tested and compared all datasets amongst each other individually and collectively using the same deep learning training model. The results show that adding different datasets for training worsen the accuracy of the model. One of the primary reasons is that the nature of these datasets vastly differs from one another.

## I. INTRODUCTION

Deception is the act of conveying or sharing ideas, concepts, or facts which have been altered for personal gain and advances. Such acts can range from winning a minor argument to manipulating the masses. It is often difficult to tell whether a piece of information provided is genuine or deceptive which is why it is important to use a deception detection system to check whether some crucial information is true or not. Several deception detection systems and techniques have been created to be able to detect such acts. Such systems can provide essential information in various fields whose decision making depends on precise, unaltered information such as investigations, court trials, job interviews (example shown in figure 1), etc.

One of the most commonly known forms of deception detection is using the polygraph. The polygraph detects deception by sensing and analyzing several physiological processes and variations in them e.g. heart rate. The polygraph can be used for multiple purposes including investigations. The varying purposes require different forms of input and output and have varying implications [2]. The output is in the form of charts that the examiner will use to infer a psychological state, namely, whether a certain confession was truthful or deceptive. The out of confessions can have varying results depending on the criteria for the usage e.g. crime investigations require very specific answers while employment screening asks generic questions.

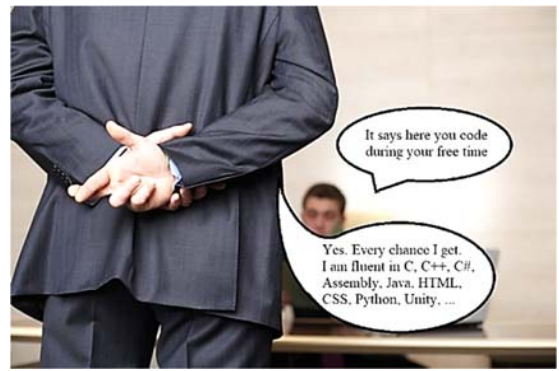

Figure 1: A scenario showing how a hobby can be anything you describe during an interview if it doesn't directly relate to the job you are applying for but can add positive points in your favor [1]

Facial expressions help showcase emotions in ways that sometimes words aren't enough. It is within this range that the premise for deception detection based on facial expressions is derived. While it is easy to detect emotions and actions like laughter, sadness, and anger, there are subtle changes that go completely unnoticed by the untrained eye. According to [3], there are 2 kinds of facial expressions. Macroexpressions are obvious to understand like anger, fear, happiness, sadness, etc. They last between 0.5 to 5 seconds. Microexpressions occur unconsciously and display a concealed emotion [4] and last less than 0.5 seconds. Amusement, anxiety, shame, embarrassment, pleasure, relief, and guilt are classified as microexpressions. While it is easy to classify macroexpressions as they occur more frequently and last longer, microexpressions go unnoticed by the untrained eye for the opposite reasons. They are developed as a result of someone trying to conceal some form of emotion or are trying to deceive someone else.

The Facial Action coding System (FACS) was created by Carl-Herman Hjortsjö [5] and later adopted and improved upon by Paul Ekman and Wallace Friesen [4]. In FACS, every facial muscle movement is encoded. Action Units (AUs) are the essential actions of distinct or group of facial muscles. It is with the combination of these AUs that a distinct facial expression (either microexpression or macroexpression) can be identified by as shown in figure 2.

| NEUTRAL | AU 1 | AU 2 | AU 4 | AU 5 |
|---|---|---|---|---|
| Eyes, brow, and cheek are relaxed. | Inner portion of the brows is raised. | Outer portion of the brows is raised. | Brows lowered and drawn together | Upper eyelids are raised. |
| AU 6 | AU 7 | AU 1+2 | AU 1+4 | AU 4+5 |
| Cheeks are raised. | Lower eyelids are raised. | Inner and outer portions of the brows are raised. | Medial portion of the brows is raised and pulled together. | Brows lowered and drawn together and upper eyelids are raised. |
| AU 1+2+4 | AU 1+2+5 | AU 1+6 | AU 6+7 | AU 1+2+5+6+7 |
| Brows are pulled together and upward. | Brows and upper eyelids are raised. | Inner portion of brows and cheeks are raised. | Lower eyelids cheeks are raised. | Brows, eyelids, and cheeks are raised. |

Figure: A selected (non-exhaustive) list of Upper Face AUs and some of their combinations 2 [9]

## II. Related Works

FACS has been effective in several studies regarding deception detection including this study [6] which extracted facial AUs with the help of openface [7]. For facial landmark detection, they used Constrained Local Neural Field (CLNF) [8]. Openface extracts over 700 features from pictures and/or videos of which 35 were related to AUs. Those 35 AU related features were then further analyzed to check their significance by checking their p-value. Of the 35, only 3 were shown to have zero significance.

This paper [10] proposes an end-to-end framework DEV, which can accurately detect **DE**ceptive **V**ideos automatically. They utilize a multimodal approach that uses audio and visuals of a video as inputs. Convolutional Neural Networks (CNN) was used for feature extraction. To capture temporal correlations between those features, they used long short-term memory (LSTM). A separate study [11] used Improved Dense Trajectory (IDT), transcript, Mel-frequency Cepstral Coefficients (MFCC), and microexpressions as features. They also used what they called Ground Truth (GT) microexpressions.

This work [12] utilized a multimodal approach based on audio, transcript, and microexpressions. In this, they classified the microexpressions using Adaboost. For their audio analysis, the proposed scheme employs the Cepstral Coefficients (CC) features and the Spectral Regression Kernel Discriminant Analysis (SRKDA) classifier. Lastly, for the text/transcript part of their multimodal approach, the sub-system they used is based on the Bag-of-N-Grams (BoNG) features and linear SVM. The decisions from all three sub-systems are fused using majority voting.

## III. Datasets

Several studies created datasets which they tested on their deception detection systems. Only a few of them are available for others to use. While it is understandable why private shouldn't be made public, the importance of sharing datasets to create deception detection systems is that they can be used to compare the effectiveness of different techniques and/or making systems that can handle a broader range of input. The impact of a database is checked based on stakes of the confessor and a description of stakes is as below:

- Low stakes: The confessor will not be impacted in any way after the recording session is over hence will not be concerned about the outcome of their answers

TABLE I. COMPARISON BETWEEN DIFFERENT DATASETS

| DATASET | NAME | NO. OF VIDEOS | FPS | STAKES | LABELS | FACIAL LABELS | AVAILABILITY |
|---------|------|---------------|-----|--------|--------|---------------|--------------|
| [13] | REAL-LIFE TRIAL DATASET | 121 | 30 | HIGH | 60 VIDEOS ARE TRUTHFUL WHILE 61 ARE DECEITFUL | MUMIN | PUBLIC |
| [14] | SILESIAN DECEPTION DATASET | 101 | 100 | LOW | EACH VIDEO CONTAINS 3 TRUTHFUL CONFESSIONS AND 7 DECEPTIVE CONFESSIONS | MANUAL | SPECIAL ACCESS |
| [15] | BAG-OF-LIES DATASET (BOL) | 325 | 30 | LOW | 163 VIDEOS ARE TRUTHFUL WHILE 162 ARE DECEITFUL | NONE | SPECIAL ACCESS |

- Medium stakes: There is a possibility that the confessor will be getting some form of advantage for their answers but not in ways that it will alter their lives too significantly e.g. a friendly game in which someone has to guess if the confessor is lying or not.
- High stakes: The life of the confessor or someone they know will be significantly altered based on their confessions e.g. a court trial which can lead to serious consequences for their or someone else's life.

Datasets that were compiled under laboratory conditions usually fall under low or medium stakes. Datasets in which the confessor is under strict or life-altering situation (e.g. court trials) fall under the category of high stakes. Table I shows the 3 datasets that we acquired along with some of their properties like FPS (Frames per Second) of the videos, the types of stakes involved, etc.

The real-life trial dataset is a collection of real-life trial case videos downloaded from YouTube. The videos were classified as either being truthful or deceitful based on the final verdict given by the judge based on evidence provided and, in some cases, based on exoneration. The dataset also provides in total 39 verbal and non-verbal cues present in each video using the MUMIN [16] coding scheme for both facial and hand movements. This is the most commonly used dataset related to deception detection mostly due to ease of access.

The Silesian deception dataset consists of 101 videos all of which were recorded at 100 FPS in a well-controlled laboratory environment and proper illumination. The videos neither have any audio nor is the transcript for their answers are provided. Instead of annotating facial AUs, the dataset provides annotation of what the author calls micro-tensions. To avoid the introduction of subtle changes in the facial expressions and blink dynamics, the subjects were not informed that the main aim of the research is facial analysis in the context of deception detection.

BOL dataset consists of video, audio, and eye gaze from 35 unique subjects collected using a carefully designed experiment. For the experiment, each subject was shown 6-10 select images and was asked to describe them deceptively or otherwise based on their choice. The frame rate for each video is 30 FPS.

IV. EXPERIMENT AND PREPROCESSING

Our objective is to use Deep Learning to determine whether a certain confession in a video is truthful or deceptive. For this, we used the trial dataset because of its nature (high stakes). Firstly, to extract AUs, we used openface [7]. Since a single frame cannot determine whether someone is lying or not, we utilized long short-term memory (LSTM) model as shown in figure 3. We analyzed all videos and optimized them by either removing some videos altogether, by editing out additional faces from them or by editing out portions that either didn't have any faces or nothing was being confessed in those sections (the confessor is silent and is listening to someone else). In the end, we had a total of 41904 video frames related to deceptive confessions and a total of 38522 video frames related to truthful confessions.

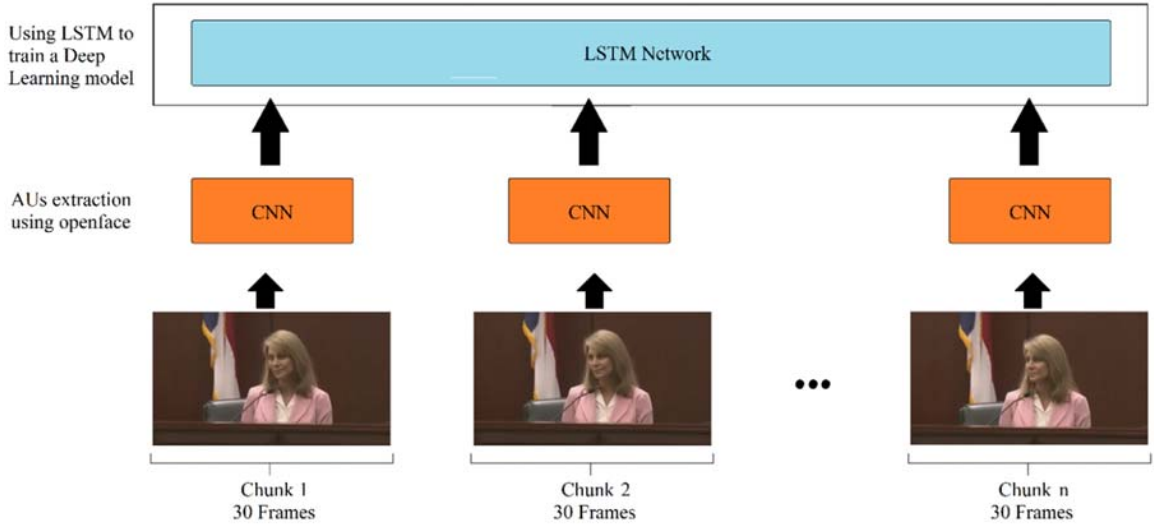

Figure 3: Training an LSTM deep learning model

To extract facial AUs from the videos, we used openface [7] which is an open-source software that utilizes deep learning to extract facial features from videos and pictures. For facial landmark detection, we used the CLNF model on the enhanced trial dataset. In total, it provided over 700 features of which 35 are related to AUs. Taking into account what [6] found, 3 of the AUs were neglected to optimize our results. Neglecting these AUs did improve the results.

Instead of using one confession as a single input, we divided them into chunks of 30 frames. This makes the size of a single chunk 30x32 (30 frames per chunk and 32 AUs). We removed some chunks (deceptive chunks were larger in quantity than truthful one) until we had a ratio of 1:1 between deceptive chunks and truthful chunks. Our experiments provided improved results from this approach. This dataset was then randomly divided by a 70:30 ratio. 70% of the samples would be used for training purposes while the remaining 30% would be used for testing.

For our deep learning approach, we utilized long short-term memory (LSTM) as shown in figure 4. While other deep learning and neural network approaches have been used for deception detection systems, in most cases they were used for the final fusion in multimodal approaches. We wanted to see the results of using some form of deep learning to see how it fairs in a visual-only or facial only approach against other approaches. The model we utilized was fairly simple. It consists of a single LSTM layer, a single dropout layer to prevent overfitting, and a dense layer with a single output (since a chunk can either be truthful or deceptive). Adding more layers to this model only worsened the results.

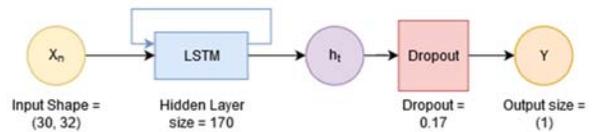

Figure 4: LSTM model

## V. RESULTS AND DISCUSSION

We processed the untrained chunks as shown in figure 5. During our initial runs, our LSTM model had a total of 4 LSTM layers, 4 dropout layers, and a single dense layer. Using this model, we were receiving a correct classification rate (CCR) of up to 85% in which we had only optimized the dataset. After applying the feature selection, maintaining a 1:1 ratio for both truthful and deceptive samples, and making the LSTM model simpler as shown in figure 4 by removing several LSTM layers, it scored a maximum CCR of 89.49% which is currently better than most visual-only and facial-only deception detection approaches. Table II shows that using deep learning and neural networks for creating deception detection systems for visual only approaches can garner great results. The program was run on python for which the code has been uploaded on GitHub [17].

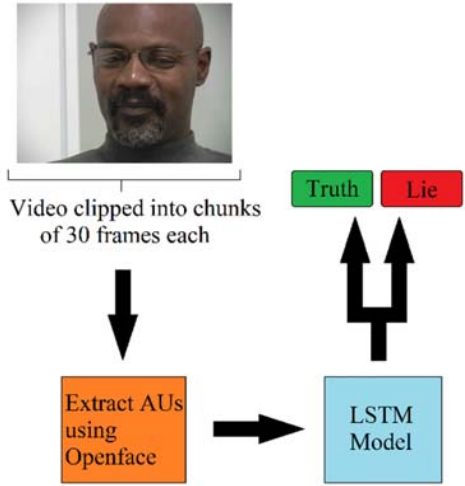

Figure 5: Processing a confession using the LSTM model

TABLE II. COMPARISON BETWEEN DIFFERENT VISUAL/FACIAL-ONLY APPROACHES

| APPROACH | METHOD | CCR |
|---|---|---|
| **CURRENT** | **FACS WITH LSTM** | **89.49%** |
| [13] | DECISION TREES AND RANDOM FOREST | 75.20% |
| [6] | SVM-RBF | 76.84% |
| [12] | ADABOOST | 88% |

While reviewing other researches and papers, it came to our attention that cross-dataset incorporation and validation has not yet been attempted by anyone else. To test this, we gained access to 3 datasets related to deception detection (the ones discussed in section III). We preprocessed them the same way we did for the trial dataset. Table II shows are our findings in terms of mixing other datasets and their impacts on individual and combined accuracies. Test accuracies were checked using data that was not used for training. This also means that if a certain dataset was not part of the training, then that whole dataset was used for accuracy testing. While we still went with maintaining a balance between truth samples and deceptive samples, the results shown omits this rule only for the Silesian dataset as using the complete dataset provided better test accuracies.

While checking the Silesian dataset, we found 2 videos to be duplicates of 2 other videos. In some videos, it was sometimes hard to decipher when a confession started and when it ended. We decided to omit these videos from testing and training samples. In the end, we had a total of 925 confessions.

TABLE III. CROSS DATASET VALIDATION RESULTS

| DATASETS USED | | | TEST ACCURACY OF INDIVIDUAL DATASETS | | |
|---|---|---|---|---|---|
| TRIAL | SILESIAN | BAG-OF-LIES | TRIAL | SILESIAN | BAG-OF-LIES |
| **YES** | NO | NO | **89.49%** | 50.86% | 48.11% |
| NO | **YES** | NO | 50.18% | **75.82%** | 49.92% |
| NO | NO | **YES** | 61.21% | 43.00% | **67.11%** |
| **YES** | **YES** | NO | **73.96%** | **70.92%** | 48.51% |
| **YES** | NO | **YES** | **77.72%** | 43.84% | **63.42%** |
| NO | **YES** | **YES** | 57.00% | **70.60%** | **64.12%** |
| **YES** | **YES** | **YES** | **72.15%** | **69.38%** | **62.37%** |

As the results show in table III, in every case, the test accuracy has worsened when the LSTM model is trained using 2 or more datasets. It most likely has something to do with the difference of nature between each dataset. The trail dataset contains examples of people in high stakes situations while the other 2 datasets were recorded in laboratory conditions. While the laboratory conditions did ensure that the confessors were unaware of the reason behind the recordings (thus ensuring that they weren't acting about lying but were genuinely lying), their future lives were not dependent on those confessions while the people in the trail dataset consisted of people whose future lives depended on their truthful or deceitful confessions. There is also the problem of the environment. The Silesian and BOL datasets were set in laboratory conditions which did ensure that the camera recordings were at optimal conditions (only the confessor present, confessor looking directly at the camera, etc.). This is not the case for the trail dataset. The Silesian dataset was also recorded at 100 FPS which is more than 3 times the FPS of the other 2 datasets.

## VI. Conclusion

In this study, we showcased several new elements to deception detecting systems that were previously untested or were used separately from other techniques. We showcased techniques like chunking, equal ratio, and deep learning as being important for increasing the effectiveness of the deception detection system. With the aid of LSTM, we not only showed that using deep learning for the development of automatic deception detection systems is viable, but it also outclasses several other approaches. The next step for improvement would be to use it for a multimodal approach as it has shown to improve results as compared to visual-only approaches. Better results can also be obtained if the accuracy of openface or any other FACS feature extractor is also increased. For better training results, the current high stakes dataset should also be updated with more confessions and removing sections that do not contain any form of confessions.

## VII. Conflict of Interest Statement

The authors have no conflicts of interest to declare. All co-authors have seen and agreed with the contents of the manuscript and there is no financial interest to report. We certify that the submission is original work and is not under review at any other publication.